\documentclass[letterpaper, 10 pt, conference]{ieeeconf} %

\IEEEoverridecommandlockouts                              %

\usepackage{amsmath} %
\usepackage{amssymb}  %
\usepackage{amsfonts}       %
\usepackage{bm}
\usepackage{cite}
\usepackage{url}
\usepackage{siunitx} %
\usepackage{subcaption}
\usepackage{graphicx}
\usepackage{footnote}

\usepackage{color,soul}

\usepackage{balance}
\usepackage{booktabs}
\usepackage{multirow}
\usepackage{xcolor}
\usepackage[normalem]{ulem}   %

\newcommand{\ra}[1]{\renewcommand{\arraystretch}{#1}}
\title{\LARGE \bf
  Learning Contact Dynamics through Touching: Action-conditional
  Graph Neural Networks for Robotic Peg Insertion
}
\author{Zongyao Yi\authorrefmark{1}, Joachim
  Hertzberg\authorrefmark{2}\authorrefmark{1} and
  Martin Atzmueller\authorrefmark{2}\authorrefmark{1}%
  \thanks{\authorrefmark{1}German Research Center for Artificial
    Intelligence, DFKI Niedersachsen, Osnabrück, Germany
    {\tt\small zongyao.yi@dfki.de}
  }%
  \thanks{
    \authorrefmark{2}Osnabrück University, Institute of Computer
    Science, Osnabrück, Germany
  }%
}

\begin{document}
\bstctlcite{BSTcontrol}

\maketitle
\thispagestyle{empty}
\pagestyle{empty}

\IEEEpeerreviewmaketitle

\begin{abstract}
  We present a learnable physics-based model that predicts motion of
  the robot end effector and
  reaction force-torque in contact-rich
  manipulation. The model represents the end effector and the
  environment as interacting
  meshes in a graph structure, and conditions its prediction
  explicitly on the applied control
  input. It predicts object-level pose update directly, while the
  reaction torque emerges from a per-vertex force field. Training is
  self-supervised using only joint encoder and
  force-torque data while the robot is randomly touching the
  environment without task context.
  In simulation, our model transfers to peg insertion with unseen
  concave geometry,
  where an MPC agent using it reaches up to 98\% success rate, and
  after fine-tuning on
  self-collected data matches an agent planning with the ground truth
  dynamics at the tightest
  1$\,$mm clearance. In the real world,
  it outperforms the system-identified MuJoCo model by 45\% in
  position and by 74\% and 63\%
  in force and torque error.
  \url{sites.google.com/view/act-fignet}.
\end{abstract}

\section{INTRODUCTION}
\label{sec:introduction}

Precise assembly tasks like peg insertion require robots
to operate under constant contact with rigid and unyielding
environment, such as a fixed
electrical socket. Unlike
interaction with movable objects, these environments are non-compliant and
pose hard geometric constraints on the robot's motion. This makes
action-conditional prediction of motion and forces non-trivial, as violating
these constraints can cause excessive contact
forces and potentially damage the robot. Analytical physics
simulators\cite{conf/iros/TodorovET12,coumans2021} can be identified
to reproduce end effector motion reasonably well, but the
point-contact approximations they rely on limit the fidelity of the
predicted contact forces once surface contacts dominate.
Recent work has
shown that Graph Neural Networks (GNNs) are more capable of approximating
real-world contact dynamics
\cite{conf/corl/AllenLRSSBP22,conf/iclr/AllenRL0SBP23}. However,
existing GNN-based approaches focus on current state to next state
prediction; the effects of
control inputs are not explicitly modeled. Moreover, these methods
cannot predict contact
forces and require privileged training data, e.g., fully observed trajectories
of free falling objects, which is hard to obtain in the real-world.

In this paper, we present a graph-based forward and observation model
that predicts both the end effector motion and reaction wrench
induced by environmental contacts and applied control input. The
model builds a graph over the tool and environment meshes, with
control input encoded as an action-conditional graph layer. Contacts
are encoded as face-face edges following the formulation
of~\cite{conf/iclr/AllenRL0SBP23}. The graph construction is
object-centric with all features represented in the body frame, such
that the model directly predicts object-level twist instead of
relying on vertex-cloud matching. For the wrench readout, our
framework introduces a per-vertex force field decoder, making
reaction torque an emergent quantity. We refer to the resulting model
as Act-FIGNet. It not only learns the smooth dynamics of the
robot end effector in the absence of contacts, but also the
discontinuities when contacts
engage. In simulation, we deploy the learned model with
model-predictive control on a challenging peg insertion task and
achieve up to 98\% success zero-shot. At the tightest 1$\,$mm
clearance, fine-tuning on self-collected data brings it on par with
an agent planning with the ground truth dynamics. In the real-world
experiment, our
model outperforms the system-identified MuJoCo baseline in both
motion and wrench prediction. In summary, this paper makes the
following contributions:
\begin{itemize}
  \item An action-conditional predictive model serving both as a
    forward dynamics model
    for model-predictive control and as an observation model for F/T
    reading, trained
    self-supervised from proprioceptive and wrench data;
  \item A body-frame graph construction invariant to global scene
    rotation and predicts the object-level pose update directly,
    rather than recovering it from vertex cloud matching;
  \item A coupled wrench readout, in which the reaction torque
    emerges from the predicted
    per-vertex force field rather than a separate regression target.
\end{itemize}
\section{RELATED WORK}
\label{sec:related_work}
\subsection{GNN-based physics simulators} Learned graph-based
dynamics were first applied to control tasks with particle and link
representations~\cite{conf/icml/Sanchez-Gonzalez18,conf/icra/Li0ZT0T19},
which model interacting neighbors and long-range propagation but not
contacts between complex geometries, and were demonstrated on simple
scenes, such as swinging pendulum and box pushing. Mesh-based
models~\cite{conf/iclr/PfaffFSB21, conf/corl/AllenLRSSBP22,
conf/iclr/AllenRL0SBP23} resolve contacts at state-of-the-art
accuracy, but are trained on passively observed rigid-body motion and
not conditioned on control input. Our model retains the mesh-based
contact resolution, but also conditions its prediction on the applied
control and is trained from robot proprioceptive and wrench data alone.

\subsection{Contact force prediction} Precise contact force
prediction helps to avoid
hazardous robot actions, and provides a discriminative observation
model for filtering-based state estimation\cite{conf/iros/Pankert023,
  journals/corr/abs-2505-19215, journals/ijrr/GadeyneLB05,
journals/corr/abs-2409-17470}. However, work on learnable contact-
and action-conditional force prediction remains scarce.
A binary force observation model is used
in~\cite{conf/iros/Pankert023}, relying on
a predefined threshold and conditioned on the contact state only.
In\cite{journals/corr/abs-2312-09190}, a
linear model predicts the F/T readings from robot's configuration and
applied control, encoding contacts implicitly, which limits transfer
to unseen connector geometries.
A particle filter coupled with an analytical simulator jointly estimates
contact forces, object poses, and geometries
in~\cite{journals/corr/abs-2409-17470}, assuming the bodies stay in
force equilibrium. Our approach makes no such assumption, and
predicts both smooth dynamics and discontinuous contact behaviors.
\section{METHOD}
\subsection{Problem statement} A manipulation task can be modeled as a
discrete-time partially observable Markov decision process (POMDP)
with continuous state, action and observation
space $\mathcal{S}$, $\mathcal{A}$, $\mathcal{O} $, transition
dynamics $\mathcal{T}$ and observation model $\Omega$. The goal of
the proposed model is to
(1)
approximate the transition dynamics $\mathcal{T}$ with a prediction function
$g: \mathcal{S} \times
\mathcal{A} \to \mathcal{S}$ to minimize
error $\|g(s_t, a_t) - s_{t+1}\|_2$; and (2) approximate the
observation model $\Omega$ with $h: \mathcal{S} \times
\mathcal{A} \to \mathcal{O}$, minimizing $\|h(s_t, a_t) - o_t\|_2$.
\subsection{Setup} We focus on the peg-in-hole task and assume the
robot operates under a Cartesian space controller,
e.g., impedance or force control. This lets us reduce the robot to a
floating tool body interacting with the environment
(Fig.~\ref{fig:robot_setup}e), moving in response to
the applied wrench, and hence independent of the robot's kinematic
structure. Since the controller compensates gravity, the tool
dynamics contains no world-frame preferred direction. Represented in
maximal coordinates,
the tool configuration is unconstrained except by conatcts, and it is written as
$s_t=[\bm{x}_t,
\dot{\bm{x}}_t]$, where
$\bm{x}_t \in \mathbb{R}^{7}$ is the 3-D position and 4-D
quaternion, and $\dot{\bm{x}}_t \in \mathbb{R}^{6}$ is the
linear and angular velocities. The control input is the 6-D external
wrench acting on the tool body, $a_t =[\bm{f}_t,
\bm{\tau}_t] \in \mathbb{R}^{6}$.
The prediction model takes as input a state sequence $\bm{s}^h_t=[s_t, ...,
s_{t-h+1}]$, together with $a_t$, to predict the next state $\hat{s}_{t+1}$ and
the reaction wrench from the tool body, i.e., the F/T sensor reading,
defined as the observation $\hat{o}_t =
[\hat{\bm{f}}_t, \hat{\bm{\tau}}_t]$.
\paragraph{Geometry and graph notation} The tool body and the
environment geometries are
represented with triangle meshes, each consisting of a set of
vertices $\{\bm{p}^{m_i}\}$ and a
set of faces $\{\mathcal{F}_k\}$ given as vertex-index triplets. For a body with
$\mathcal{N}$ vertices we write $\langle\bm{x}\rangle \equiv
\frac{1}{\mathcal{N}}\sum_{i}\bm{x}_i$
for the mean of a per-vertex quantity, and $\mathring{\bm{x}}_i
\equiv \bm{x}_i -
\langle\bm{x}\rangle$ for the centered quantity. A directed graph
$\mathcal{G}=(\mathcal{V},
\mathcal{E})$ carries a feature vector $\bm{v}_i$ on each node and
$\bm{e}_{i \rightarrow j}$ on
each edge.
\subsection{Pipeline}
\begin{figure*}[t]
  \centering
  \includegraphics[width=0.99\textwidth]{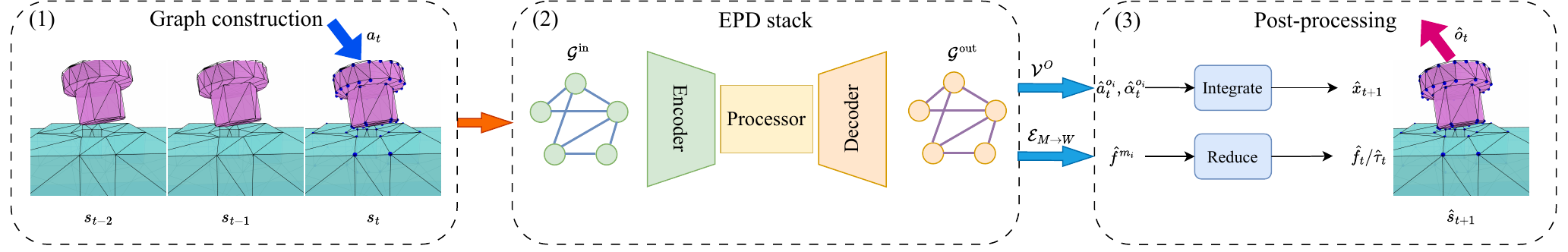}
  \caption{Model pipeline overview. Cf. text for details about
    (1) Graph construction, (2) EPD stack, and (3) Post-processing.
  }
  \label{fig:pipline_overview}
\end{figure*}

The pipeline consists of three main components: a graph constructor, the
encoder-processor-decoder (EPD) stack, and the post-processing
module. This stack follows the standard recipe of graph neural
network simulators~\cite{conf/iclr/AllenRL0SBP23,conf/iclr/PfaffFSB21}.
The pipeline first builds the input graph $\mathcal{G}^{\text{in}}$
from $\bm{s}_t^h$ and $a_t$,
given the object meshes and their static attributes, i.e., mass and
friction coefficient\footnote{Both are constant in our data and
  therefore carry no information; the fields are kept so that it
extends to scenes with varying materials.}. The EPD
stack maps it to an output graph, from which the post-processing
module reads out the predicted
next state $\hat{s}_{t+1}$ and observation $\hat{o}_t$, together
approximating $g$ and $h$.
Fig.~\ref{fig:pipline_overview} gives an overview; each component is
described below.
\subsubsection{Graph construction}

\begin{table}[tb]
  \centering
  \caption{Elements of the input graph $\mathcal{G}^{\text{in}}$.}
  \label{tab:graph_spec}
  \footnotesize
  \setlength{\tabcolsep}{3pt}
  \ra{1.2}
  \begin{tabular}{@{}ll p{4.9cm}@{}}
    \toprule
    Element &   & Features \\
    \midrule
    \multicolumn{3}{@{}l}{Nodes} \\
    $\mathcal{V}^M$ & Mesh vertex & Vertex velocities; static attributes \\
    $\mathcal{V}^O$ & Object CoM & Linear/angular velocity; static attributes \\
    $\mathcal{V}^W$ & Wrench node & Constant embedding \\
    \addlinespace
    \multicolumn{3}{@{}l}{Edges} \\
    $\mathcal{E}^{O \leftrightarrow M}$ & Object $\leftrightarrow$
    vertex & Lever arm $\bm{r}$ from the object's CoM\\
    $\mathcal{E}^{W \leftrightarrow M}$ & Wrench $\leftrightarrow$
    vertex & Coupled force field (See \eqref{eq:force_edge_feature}) \\
    $\mathcal{E}^{M \leftrightarrow M}$ & Colliding face pair &
    Face-interaction contact features \\
    \bottomrule
  \end{tabular}
\end{table}

\begin{figure}[!b]
  \centering
  \includegraphics[width=0.97\linewidth]{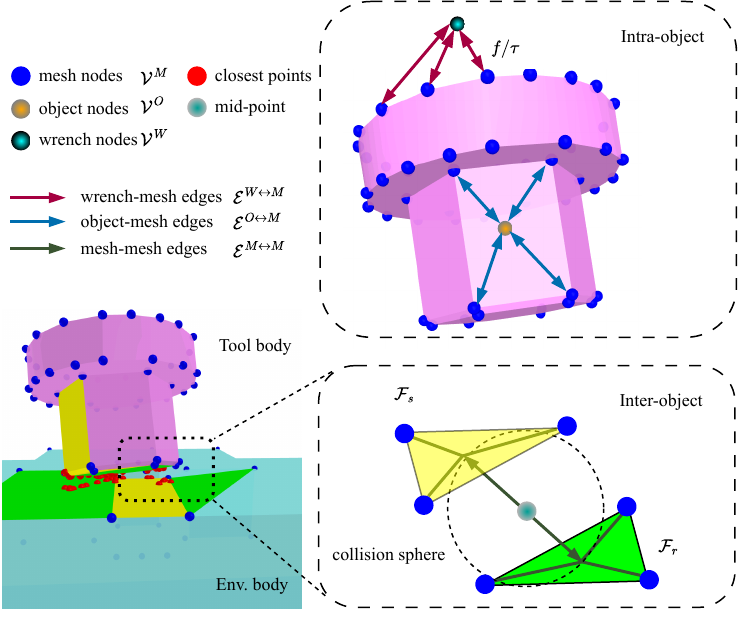}
  \caption{Action-conditional edges between the wrench nodes and the tool mesh
    nodes (top). Face interaction mesh-mesh edges are created when two faces
    ($\mathcal{F}_s$ and $\mathcal{F}_r$) fall within a predefined
  collision sphere (bottom).}
  \label{fig:graph_builder}
\end{figure}

The system is encoded in a heterogeneous graph
$\mathcal{G}^{\text{in}}$ specified in Table~\ref{tab:graph_spec}.
All features are represented in the object's body frame. The
mesh-mesh collision edge features
follow~\cite{conf/iclr/AllenRL0SBP23}, and are projected to the
receiver object's frame. Body frame representation makes our graph
construction invariant to global rotations of the scene, and hence
the prediction rotation-equivariant. Fig.~\ref{fig:graph_builder} visualizes the
graph structure. Wrench node $\mathcal{V}^W$
and the wrench-mesh edges $\mathcal{E}^{W \leftrightarrow M}$ are
introduced to encode control inputs. Per-object wrench node is
connected to the vertices belonging to the controlled object. The wrench node
is simply an anchor for the wrench-vertex edges. In our experiments,
the Cartesian controller parameters are tuned to produce stable
contact behaviors. These parameters are held fixed and not inputs to
the model, so the learned model is specific to the controller used
during data collection. The wrench node provides a natural place to
condition on controller parameters when these vary, which we leave to
future work. The edge features are the vertex-distributed wrench
command, with torque as a local equivalent force:
\begin{equation}
  \bm{f}^{m_i} = \bm{f}_t + \mathring{\bm{\mu}}_i
  \qquad
  \bm{\mu}_i = \frac{\bm{\tau}_t \times
  \mathring{\bm{r}}_i}{\lVert\mathring{\bm{r}}_i\rVert^{2}}
  \label{eq:force_edge_feature}
\end{equation}

Averaging the force field recovers the commanded force, since
centering $\bm{\mu}_i$ makes the torque-induced forces cancel; the
torque is recovered by the weighted inverse in
\eqref{eq:wrench_reduce}. Representing torque as a force field
distributed around the centroid means the vertex nodes only need to
represent linear quantities such as velocity and acceleration. For
static bodies, the edge features are
simply zero vectors. In the following, we
use $\bm{v}^{X,l}$ and $\bm{e}^l_{X \to Y}$ to denote node and edge
features, where $X, Y \in \{M, O, W\}$ are placeholders for different node
types, and the superscript ${(\cdot)}^l$
indicates the message-passing layer.
\subsubsection{Encoder-Processor-Decoder}
The per-type encoders map each node and edge feature vector to a
latent embedding,
giving the first layer embeddings $\bm{v}^{X,1}_i$ and $\bm{e}^1_{X_s \to Y_r}$.
The processor consists of $N=6$ message-passing layers, and each layer performs
edge feature updates followed by message aggregation and node feature updates:
\begin{align}
  \bm{e}^{l+1}_{X_s \to Y_r} &=
  \phi^{\text{proc}_{l}}_{\mathcal{E}^{X \to
  Y}}(\bigl[\bm{e}^{l}_{X_s \to Y_r}, \bm{v}^{X, l}_{s}, \bm{v}^{Y,
  l}_{r}\bigr]) \label{eq:edge_update} \\[1ex]
  \bm{v}^{Y, l+1}_r &=
  \phi^{\text{proc}_{l}}_{\mathcal{V}^Y}(\bigl[\bm{v}^{Y, l}_r,
  \sum_s{\bm{e}^{l+1}_{X_{s} \to Y_r}}\bigr]) \label{eq:node_update}
\end{align}
The node/edge embeddings in the final graph are then passed to the
per-type decoders. The object node embeddings are decoded into 6-D
translational and rotational accelerations, and per-vertex reaction
force field from the wrench-mesh edge latents:
\begin{gather}
  \bigl[\hat{\bm{a}}^{o_i}_t,
  \hat{\bm{\alpha}}^{o_i}_t\bigr]=\phi^{\text{dec}}_{\mathcal{V}^O}(\bm{v}^{O,
  N}_i) \\[1ex]
  \hat{\bm{f}}^{m_i} = \phi^{\text{dec}}_{\mathcal{E}^{M \to
  W}}(\bm{e}^{N}_{m_i \to w})
\end{gather}
All functions in this module are implemented as 2-layer MLPs with
latent dimension of 64. Because the object nodes receive information
only from the mesh nodes, the object-mesh edge functions need to
capture the first-moment relations of the form $\bm{r}\times\bm{v}$
and $\bm{r}\times\bm{f}$. We therefore add an antisymmetric bilinear
channel to the edge functions~\cite{conf/iclr/KimOLKHZ17,
conf/iclr/BrandstetterBWG23}. Its contribution is quantified in
Table~\ref{tab:bilinear}.
\subsubsection{Post-processing}
The predicted accelerations\footnote{Accelerations are scaled by
  $\Delta t^2$. $\oplus$ and $\ominus$ denote composition and
difference of rotations in the world frame.} are first transformed to
the world frame with the previous body frame then the object's next
position is calculated as
$\hat{\bm{p}}_{t+1}^{o_i}=\hat{\bm{a}}_t^{o_i} + 2\bm{p}_t^{o_i} -
\bm{p}_{t-1}^{o_i}$, orientation $\hat{\bm{q}}_{t+1}^{o_i} =
\hat{\bm{\alpha}}_t^{o_i} \oplus \big[\big(\bm{q}_t^{o_i} \ominus
\bm{q}_{t-1}^{o_i}\big) \oplus \bm{q}_t^{o_i}\big]$. Per-vertex force
readouts $\{\hat{\bm{f}}^{m_i}\}$ are reduced to body wrench:
\begin{gather}
  \label{eq:wrench_reduce}
  \hat{\bm{f}}_t    = \langle \hat{\bm{f}}^{m} \rangle
  \qquad
  \hat{\bm{\tau}}_t = \bm{G}^{-1}\big\langle \mathring{\bm{r}} \times
  \hat{\bm{f}}^{m} \big\rangle
\end{gather}
where $\bm{G} = \langle\bm{P}\rangle$, with $\bm{P}_i = \bm{I} -
\bm{u}_i\bm{u}_i^\top$
and $\bm{u}_i = \mathring{\bm{r}}_i/\lVert\mathring{\bm{r}}_i\rVert$.
This is simply the inverse of \eqref{eq:force_edge_feature}, and
makes reaction torque an emergent quantity of the force field
prediction rather than a separate regression target, such that force
and torque are physically coupled by
construction\footnote{$\mathring{\bm{r}}$ is taken about the vertex
  centroid, distinct from the CoM lever arm $\bm{r}$ used as the edge
  feature. $\bm{G}$ depends only on the lever-arm directions and is
  singular only if they are collinear. Since lever arms are in body
frame, $\bm{G}^{-1}$ only needs to be computed once for each object.}.

\subsection{Training}
\paragraph{Dataset} The EPD stack is trained with the dataset of the form
$\{(\bm{s}_k^h, a_k, o_k, s_{k+1})_{k=1...D}\}$, where $\bm{s}_k^h$
is the $h$-frame state
sequence\footnote{We use $h=3$ in experiments. For sensitivity to
$h$, see Section~\ref{sec:model_analysis}.}, $a_k$ is the applied
action, $o_k$ the observation, and $s_{k+1}$ is the target next state.
The labels are the
next frame's object pose, force, and torque readings. Input sequences
are injected with Gaussian noise for rollout stability. All
inputs/outputs are normalized during training.
\paragraph{Training target and loss calculation}Our network regresses
an object-level pose
update, while the predicted wrench is obtained by reducing the
per-vertex force head through
\eqref{eq:wrench_reduce}. The object pose target is the acceleration
computed from the target
and two preceding frames. Because the reduction maps $\mathcal{N}$
per-vertex forces
onto a 6-D wrench, supervising the object-level outputs alone leaves
the per-vertex predictions
unconstrained, so we add auxiliary per-vertex position and force
losses. The per-vertex position
target is formed the same as for the object head, and the per-vertex
force target follows \eqref{eq:force_edge_feature} with the control
input $[\bm{f}_t, \bm{\tau}_t]$ replaced by the
sensor reading $[\bm{f}^{\text{sensor}}_t, \bm{\tau}^{\text{sensor}}_t]$.

The training loss sums object-level pose and wrench errors with the
vertex-level position and force errors:
\begin{equation}
  \mathcal{L} = \mathcal{L}^{O}_{\text{pos}} + \mathcal{L}^{O}_{\text{rot}}
  + \mathcal{L}^{O}_{\text{wrench}}
  + \lambda\big(\mathcal{L}^{M}_{\text{pos}} +
  \mathcal{L}^{M}_{\text{force}}\big)
\end{equation}
where superscripts $O$ and $M$ denote object- and vertex-level terms,
all computed as MSE on normalized quantities, and $\lambda = 0.1$
weights the auxiliary vertex terms.
\section{EXPERIMENT}
\label{sec:experiment}
The experiments focus on the model's action-related performance
within the peg-in-hole context. In simulation, we train the model on
MuJoCo data and compare the peg insertion success rate of an MPC
agent using our model against agents using the baseline dynamics
models. On the real-world hardware, we train on data collected from
our setup and evaluate the model in both roles: as a forward dynamics
model, through multi-step prediction of the end effector trajectory;
and as an observation model, through one-step prediction of the F/T
readings from a known state.

\subsection{Simulation experiment}
\label{sec:simulation_experiment}
\begin{figure}[tb]
  \centering
  \includegraphics[width=0.99\linewidth]{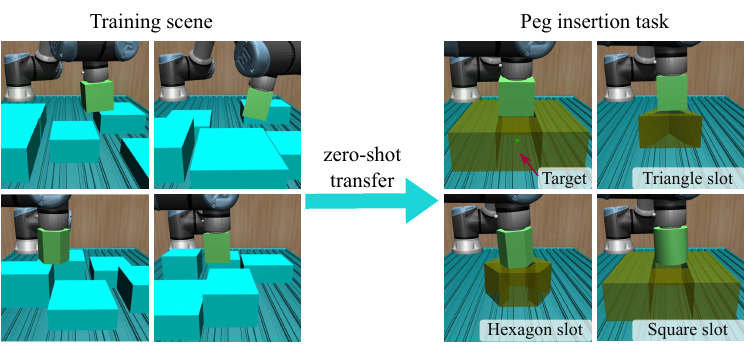}
  \caption{The training scene consists of a UR10e robot with three random tool
    shapes
    and three to six box obstacles (left). Peg insertion task scene
    with different tools and
  their matching slot geometries (right). }
  \label{fig:robot_train_task}
\end{figure}

\paragraph{Setup}
We build a MuJoCo simulation that mirrors our real-world
setup: a UR10e robot with a primitive tool mounted on its end
effector (see Fig.~\ref{fig:robot_train_task}). During data collection,
tool shapes are randomly chosen from three primitives (triangle, square,
and hexagon). To diversify the training scene, three to six box obstacles of
varying dimensions
are placed and fixed on the table. The robot is driven by a
closed-loop Cartesian force
controller\footnote{Controller reuses utilities of
robosuite\cite{journals/corr/abs-2009-12293}.}, which
accepts task-space force and torque as
control targets; control
signals are random splines interspersed with long press holds
generated at the start of each episode. Each episode
comprises 300 control steps, each subdivided into 25
simulation sub-steps. We sample 300k steps for training. The model is
trained for 30 epochs with a batch size of 256 and an initial
learning rate of $10^{-3}$, decayed by a factor of 0.1.

The peg insertion task reuses the same robot and controller, but adds a
fourth circle tool shape, which is absent from the training set. A matching
slot with depth 10\,cm, and a nominal clearance of 1\,mm, 2\,mm or
5\,mm is fixed on the table (see Fig.~\ref{fig:robot_train_task}), so
the tool must be aligned within 1 to 5\% of the insertion depth;
the training scene contains only convex box obstacles, so the concave
slot geometry is new to the model for all tool shapes. The circle
tool is paired with a
square slot. The initial tool pose is randomly sampled near the slot.
The reward function is defined on the end state: the tip-target displacement
$\Delta\bm{p}_T$ with axial component $\Delta z_T$, and the peg tilt angle
$\Delta\theta_T$. A final reward
of 10 is added when the axial error falls below a threshold $\epsilon_z$
(2$\,\mathrm{mm}$):
\begin{equation*}
  r_T = 10\,\mathbb{I}_{\Delta z_T \leqslant \epsilon_z}
  + \rho(\Delta\bm{p}_T) - \tau(\Delta\theta_T)
\end{equation*}
where the position term $\rho$ rewards axial progress gated by
lateral alignment and the orientation term $\tau$ penalizes only
tilts beyond the tolerance.
The agent uses a sampling-based method
iCEM\cite{journals/corr/abs-2008-06389} to optimize the trajectory.
Since the specifics of the MPC algorithm
are not the focus of this work, agent parameters for all models are
empirically selected\footnote{Reward function and MPC details in
Appendix~\ref{app:mpc_details}}.
\paragraph{Baselines and model variants}
As an unstructured baseline, we train an ensemble dynamics model on the
same data. The ensemble consists of 5 MLPs, each with 3 layers and a hidden
dimension of 256, which take a sequence of the peg's poses,
velocities and control inputs to
predict the next velocity; the ensemble's parameter count is of the
same order as our model.

For the most challenging 1mm tasks, we also add an oracle agent that
plans with the ground
truth dynamics to separate the effect of model error from the
limitation of the planner; and two variants
of our model: \textit{Act-FIGNet-F} is fine-tuned for around 500
optimization steps with a batch size of 64
on the base model's self-collected 120 insertion trajectories in the 5mm task.
\textit{Act-FIGNet-A} is trained from scratch on the combined dataset,
otherwise following the same schedule as the base model. We also fine
tune the MLP-Ensemble with the same recipe (\textit{MLP-Ensemble-F}).

Each MPC agent performs the tasks over 120 episodes, 30 per tool shape with
different seeds. Episodes last up to 300 control steps for 1\,mm and 2\,mm, and
150 steps for the easier 5\,mm tasks.

\begin{figure}[t]
  \centering
  \includegraphics[width=0.99\linewidth]{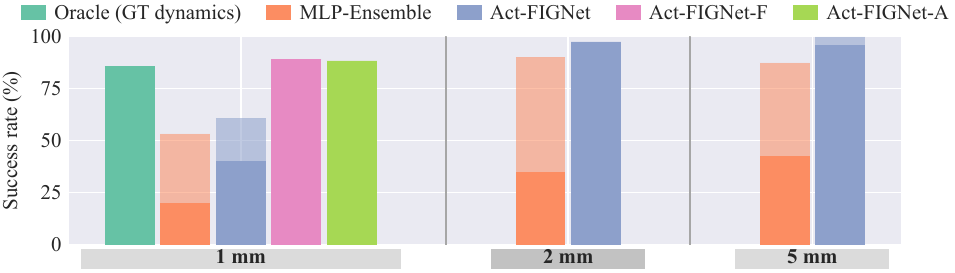}
  \caption{Success rates on the peg insertion task with different
    clearances. Solid bars are completed
    insertions, light bars in-bore trials. The oracle agent and
    the two model variants are evaluated on the 1$\,$mm task only. The
    \textit{MLP-Ensemble-F} baseline achieves 0\% and hence is omitted
  from the figure.}
  \label{fig:success_rate_comparison}
\end{figure}

\paragraph{Results}
A trial is counted successful when the tool tip touches the bottom
($\Delta z_T<2\,\mathrm{mm}$). We also count the case in which the tip has
entered the slot ($\Delta z_T<d_\text{depth}$, in-bore), with
$d_\text{depth}$ the slot's depth. The gap between the two levels
measures axial progress after alignment.
As shown in
Fig.~\ref{fig:success_rate_comparison}, the agent using the learned
model achieves a success rate of 96 to 98\% in the 2mm and
5mm tasks across all tool shapes, while the MLP baseline reaches only
35--42\% despite entering the bore in roughly 90\% of the trials.
This demonstrates our model's
generalization to concave slot geometries and to closed-loop MPC actions,
neither of which appear in the training data.

\begin{figure}[b]
  \centering
  \includegraphics[width=0.9\linewidth]{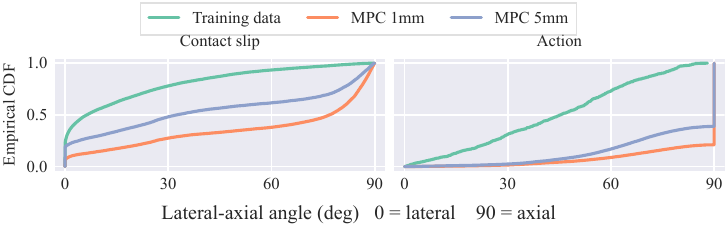}
  \caption{Lateral-axial angle for contact slip (left) and action
    (right) in the training data and the 1$\,$mm/5$\,$mm insertion
    tasks. Angles are measured from the plane orthogonal to the tool
    axis; contact slip from tangential sliding velocity, action from
    the commanded force. The 5$\,$mm data come from the base model, the
    1$\,$mm data from \textit{Act-FIGNet-F}, since the base model
  rarely completes that task.}
  \label{fig:contact_regime}
\end{figure}

In the most difficult 1mm tasks, the MLP baseline succeeds in 20\% of
the trials and enters the bore in 53\%; the base model succeeds in
40\% and enters the
bore in 60\%, meaning that a third of its aligned trials stall during
axial motion.  The progressively shifted contact regime may better
explain the stall and success rate drop than action distribution shift, as the
actions across all three clearances share very similar profiles. This
is evident from Fig.~\ref{fig:contact_regime}: sliding in the
training data is almost entirely lateral, with 6\% contact
slip above \ang{60}, compared to 38\% in the 5mm
task and 63\% at 1mm; actions differ far less between the two tasks, both concentrated near the axial direction and far from the training distribution. Against the oracle agent's 86\% at
1mm, the base model's further drop to 40\% is due to model error.
Fine-tuning and data augmentation with the base model's own 5mm
insertion rollouts close the contact regime gap. Both variants
achieve 88--89\% success matching the oracle within noise on the
unseen 1\,mm clearance. The same fine-tuning recipe does not transfer
to the MLP baseline, whose success rate drops to 0\%
(Table~\ref{tab:sim_accuracy_success}).


\paragraph{Prediction accuracy and success rate}
Table~\ref{tab:sim_accuracy_success} relates prediction error to
insertion success. Within the Act-FIGNet family, low rollout RMSE
relates to higher success rates. Across families the relation breaks
down. \textit{MLP-Ensemble-F} attains a position error below
\textit{Act-FIGNet-F}, yet never completes an insertion. Low rollout
error alone is therefore not sufficient.
\begin{table}[!tb]
  \centering
  \caption{Prediction accuracy and success rate on the 1$\,$mm
    insertion tasks. Rollout errors are 12-step (planning horizon) RMSE
    on 4k segments from the 1mm episodes, sampled by the
    \textit{Act-FIGNet-F} agent; the oracle agent plans with the ground
  truth dynamics and has no prediction error.}
  \label{tab:sim_accuracy_success}
  \footnotesize
  \setlength{\tabcolsep}{4pt}
  \ra{1.2}
  \begin{tabular}{@{}l
      S[table-format=2.2, round-mode=places, round-precision=2]
      S[table-format=2.2, round-mode=places, round-precision=2]
    S[table-format=3.0]@{}}
    \toprule
    {Model} & {Position (mm)} & {Orientation (deg)} & {Success (\%)} \\
    \midrule
    MLP-Ensemble           & 0.6283 & 0.1394 & 20 \\
    \textit{MLP-Ensemble-F}& 0.1158 & 0.0361 & 0 \\
    Act-FIGNet             & 0.3128 & 0.1553 & 40 \\
    \textit{Act-FIGNet-F}  & 0.1312 & 0.0885 & 89 \\
    \textit{Act-FIGNet-A}  & 0.0827 & 0.0636 & 88 \\
    \midrule
    Oracle (GT dynamics)   & {--} & {--} & 86 \\
    \bottomrule
  \end{tabular}
\end{table}

\subsection{Real world experiment}
\label{sec:real_world_experiment}
\begin{figure}[tb]
  \centering
  \includegraphics[width=0.99\linewidth]{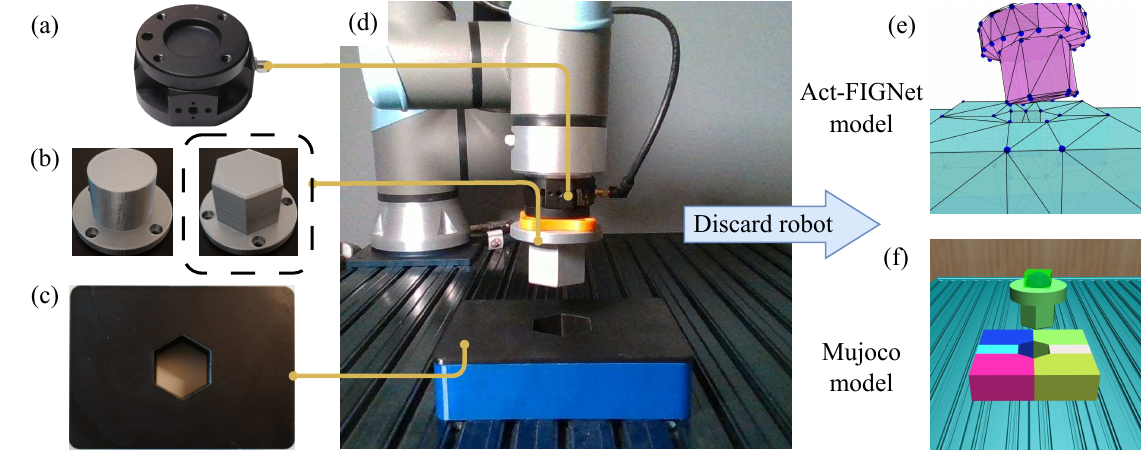}
  \caption{(a) Bota SenseOne F/T sensor; (b) 3D printed tools; (c)
    Hexagonal slot with small
    clearance; (d) Setup for the real-world experiment with UR10e; (e)
    The system contains only the tool and world body represented with
    triangle meshes; (f) In MuJoCo, the tool is modeled as a floating
  free body with 3 slide joints and a ball joint. }
  \label{fig:robot_setup}
\end{figure}

\begin{figure*}[tb]
  \centering
  \includegraphics[width=0.99\linewidth]{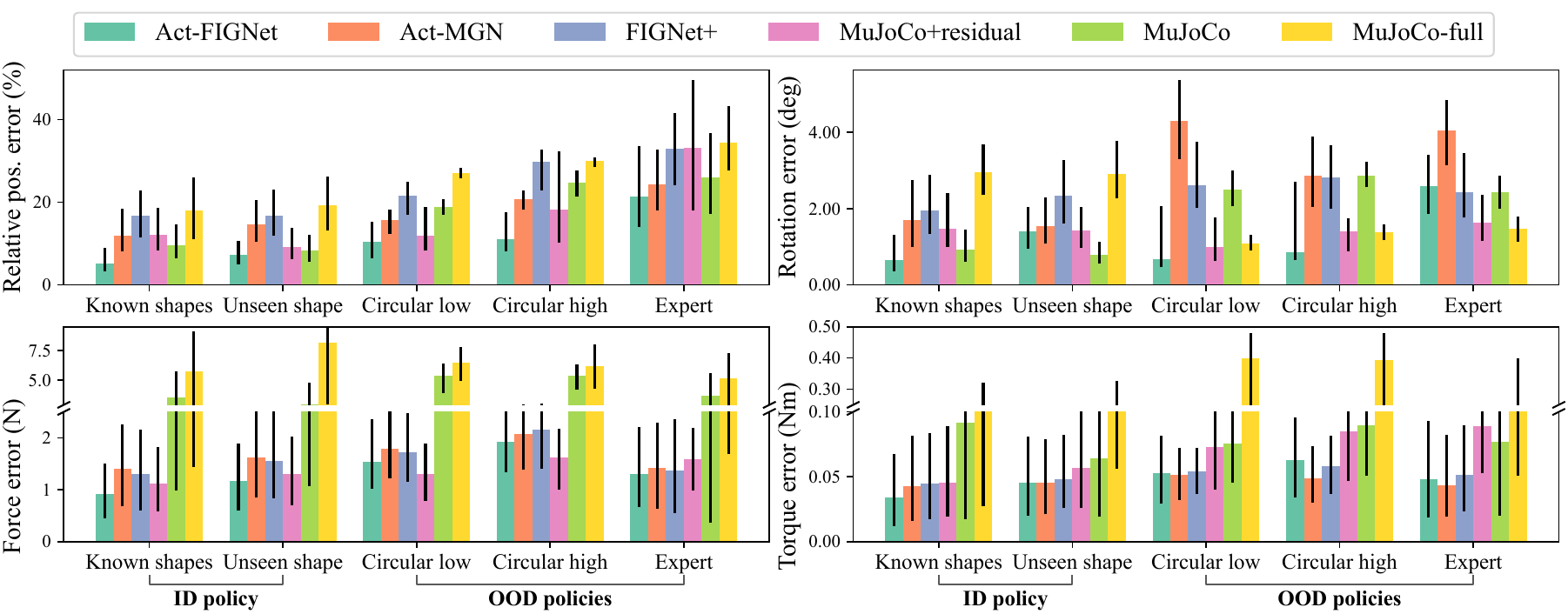}
  \caption{Prediction errors under different policies. Expert policy
    performs peg
    insertion while the circle policy controls the tool to follow
    circular paths around the slot's edge at two velocities (low and
    high). Absolute rotational error is reported due to small
    overall orientation changes. Errors under the random (ID) policy
  are included for comparison.}

  \label{fig:ood_action_error}
\end{figure*}

In this section, we test on the hardware whether our model can also
capture real-world contact dynamics by predicting motion and
force-torque readings precisely.
\paragraph{Data collection and training}
Real-world experiments are performed on a UR10e robot fitted with a
high-precision force-torque sensor and a custom 3D-printed tool
mounted on its end
effector (Fig.~\ref{fig:robot_setup}). For this experiment, we collect data
with two tool geometries, i.e., hexagonal
and circular, against a complementary hexagonal slot with a nominal
clearance of 1\,mm fixed on the test table
(see Fig.~\ref{fig:robot_setup}). The robot is
controlled by Cartesian force controller\cite{conf/iros/ScherzingerRD17}
at 500Hz, whose control target changes at 10Hz and is randomly
generated as in the simulation. At the start of each episode, the tool pose is
reset randomly around the slot geometry.
We record the control inputs as well as
the end effector pose and F/T readings. In
total, 450k steps are sampled, of which 400k for training and 50k for
testing. For evaluation only, we additionally record data with a
square tool, which is unseen
during training and tests generalization to a new geometry. The model is trained
for 50 epochs and otherwise same as in the simulation experiment.
\paragraph{Baselines} We select MuJoCo as the baseline due to its
high-fidelity contact dynamics, reported to outperform PyBullet on
real-world data~\cite{conf/iclr/AllenRL0SBP23}. In MuJoCo, we
implemented a similar closed-loop Cartesian force controller as the
real-world robot. But the control interface differs: real robot
operates under joint position control, whereas in MuJoCo under torque
control, so the simulated robot has to reproduce the arm's actuation
as well as contact behaviors.
We additionally set up a simplified MuJoCo baseline in which the
robot structure is
discarded and only the tool is kept, modeled as a rigid body with
three slide joints
and a ball joint with an F/T sensor attached
(Fig.~\ref{fig:robot_setup}f). Control
inputs are applied to the tool body directly as external wrench. System
parameters of both baselines are tuned to minimize the MSE between
simulated and ground-truth trajectories and F/T readings over 20k steps of
real-world data (Appendix~\ref{app:baseline_details}). After identification, the
simplified model is the more accurate of the two, except for rotation
under the OOD
policies. We therefore report it as the MuJoCo baseline in
Table~\ref{tab:accuracy_compare}, and the full robot model only in
Fig.~\ref{fig:ood_action_error}.

To test whether correcting an identified analytical simulator is
sufficient, or the contact dynamics has to be learned end-to-end, we
further augment the simplified MuJoCo model with a learned residual,
a 3-layer MLP with hidden dimension 128, trained on the same data as our model.

As GNN baselines, we evaluate the action-augmented versions of FIGNet
and MeshGraphNet (MGN)~\cite{conf/iclr/PfaffFSB21}. The FIGNet
variant (FIGNet+) concatenates control inputs to its node features
and decodes force-torque from the final layer node latents. The MGN
variant (Act-MGN) is overlaid with a similar action-conditional graph
layer as ours, but without coupled force field and decodes force and
torque directly from the wrench-mesh edges. Both baselines share the
same hyperparameters and training datasets.
\begin{figure}[tb]
  \centering
  \includegraphics[width=0.9\linewidth]{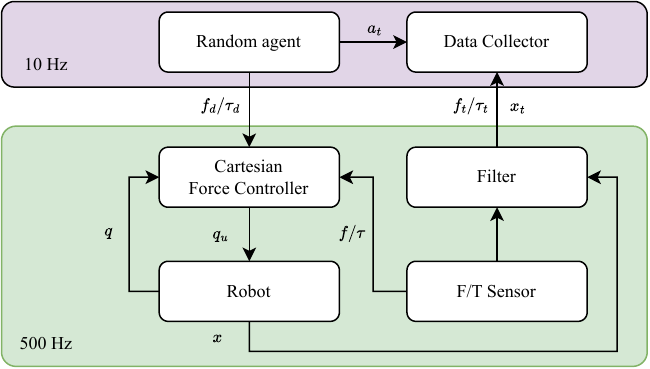}
  \caption{Control and data collection pipeline for the real-world
    experiment. The random agent outputs target wrenches
    $\bm{f}_d/\bm{\tau}_d$ to a closed-loop force controller, and the
    tool pose $\bm{x}$ is obtained from the joint positions $\bm{q}$
    by forward kinematics. Since data collection and the control loop
    run at different frequencies, the recorded F/T readings
    $\bm{f}_t/\bm{\tau}_t$ and pose $\bm{x}_t$ are filtered for
    synchronization and smoothing.
  }
  \label{fig:robot_ctrl_pipeline}
\end{figure}
\paragraph{Accuracy evaluation}We use 100-step (10\,s) rollout RMSE
over 4k test trajectories to evaluate our approach as a forward
dynamics model; and use one-step F/T error over 40k transitions to
evaluate it as an observation model, predicting from a known state
the force-torque sensor readings. Relative rollout errors are
obtained by normalizing with the accumulated state change of the ground truth.

\begin{figure}[tb]
  \centering
  \includegraphics[width=0.99\linewidth]{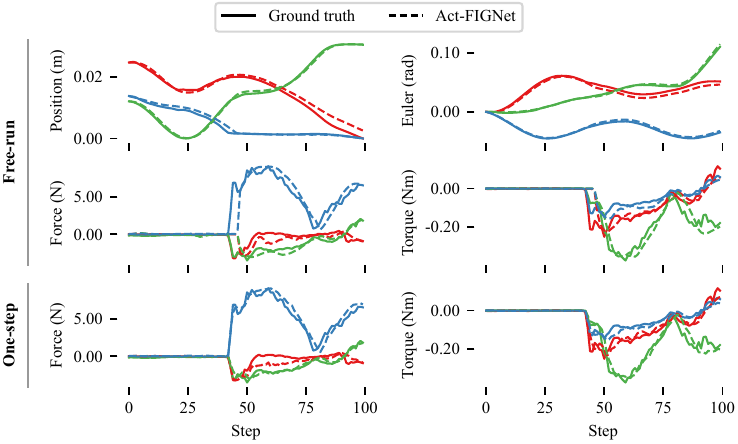}
  \caption{Prediction on a 100-step real-world test sequence.
    Top: tool position and orientation, with values shifted for visualization.
    Center: predicted wrench in free-run. Bottom: wrench predicted
    one step ahead from a known state. The x,
  y, z components are shown in red, green, and blue.}
  \label{fig:prediction_random_compare}
\end{figure}

\begin{figure}[tb]
  \centering
  \includegraphics[width=0.99\linewidth]{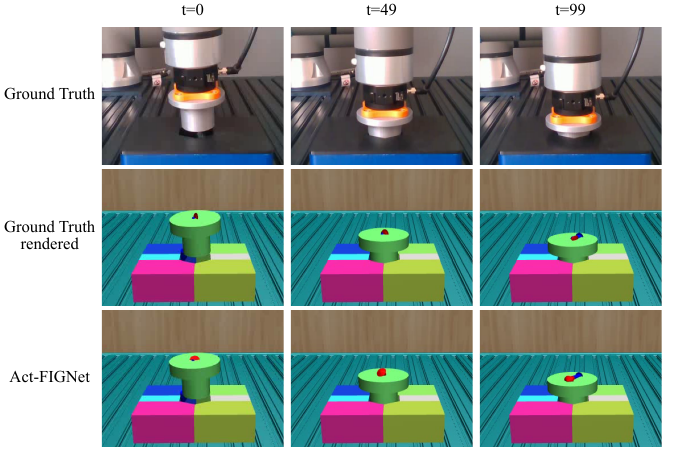}
  \caption{Predicted rollout of an example peg insertion trajectory.
  The robot in the rendering is omitted for visualization.}
  \label{fig:prediction_snapshots}
\end{figure}

Table~\ref{tab:accuracy_compare} summarizes the quantitative results.
Our model achieves 100-step RMSE of 2.87\,mm (5.25\%) in position and
0.64\,deg in orientation, with one-step F/T errors of
0.918$\,\mathrm{N}$ and 0.034$\,\mathrm{Nm}$. Against the identified
MuJoCo model these are 45\% and 30\% lower in motion, and 74\% and
63\% lower in force and torque.
Every baseline trades one metric against the other: MuJoCo predicts
motion better than the three learned baselines but is the least
accurate in F/T, while the three at least halve MuJoCo's F/T error
but predict motion less accurately.
Our model is the most accurate in all four metrics;
Fig.~\ref{fig:prediction_random_compare}
and~\ref{fig:prediction_snapshots} illustrate this accuracy.
\paragraph{Action distribution shift} Sampling-based planning queries
the dynamics model with actions far outside the training
distribution, as the simulation experiments show. We test the same
shift on hardware with three OOD control policies: a hand-coded peg
insertion (expert) controller and a circular search controller at two
velocities that moves the tools around the slot's perimeter. The
circular policies retain the lateral sliding contact of the training
data, while expert policy adds axial slip, which separates the effect
of the shifted actions from that of the contact
regime. We sample 20k steps per policy.

Results are summarized in Fig.~\ref{fig:ood_action_error}, together
with an unseen square
tool evaluated under the training policy.
Our model's prediction errors remain low under the circular policies
but rise fourfold under the expert policy relative to the
in-distribution metrics. This is expected in insertion, where lateral
errors accumulated over the 100-step open-loop rollout translate into
large axial deviation, which is why all models degrade here. The
contact regime coverage gap shown in
Section~\ref{sec:simulation_experiment} may also contribute. Across
the two circular policies, position errors grow with sliding velocity for all
models. Our model retains the lowest position error under every
policy and on the unseen
tool, while the F/T differences between the learned models are
marginal under the OOD
policies. Rotation under the expert policy is an exception, where all
MuJoCo variants predict orientation more accurately.
Act-MGN, carrying our action-conditional layer, predicts position
more accurately than FIGNet+ in all conditions, while its orientation
error is higher under the OOD policies. Its F/T errors are on par
with those of FIGNet+, which suggests that the action-conditional
layer alone does not account for our wrench accuracy. Both baselines
recover the tool's orientation from the predicted vertex cloud by a
least-squares fit, which is ill-conditioned about the tool's own
axis, especially for the circular tool. Our object-level readout
predicts the twist directly and is not subject to this.

\begin{table}[tb]
  \centering
  \caption{Prediction accuracy comparison on the real-world
    random policy test set. Shown are the 100-step RMSE for position
    and orientation;
  one-step F/T prediction error.}
  \label{tab:accuracy_compare}
  \footnotesize
  \setlength{\tabcolsep}{6pt}
  \ra{1.2}  %
  \begin{tabular}{@{}l
      S[table-format=1.2, round-precision=2]
      S[table-format=1.2, round-precision=2]
      S[table-format=1.2, round-precision=2]
    S[table-format=1.3, round-precision=3]@{}}
    \toprule
    \multirow{2}{*}{Model}
    & \multicolumn{2}{c}{RMSE\,($T\!=\!100$)}
    & \multicolumn{2}{c}{One-step Error} \\
    \cmidrule(lr){2-3} \cmidrule(lr){4-5}
    & {Pos.\,(mm)} & {Ori.\,(deg)}
    & {Force\,(N)} & {Torque\,(Nm)} \\
    \midrule
    MuJoCo        & 5.2059 & 0.9231 & 3.5420 & 0.0920 \\
    MuJoCo+res    & 6.6267 & 1.4698 & 1.1178 & 0.0454 \\
    FIGNet+       & 9.0795 & 1.9538 & 1.3039 & 0.0447 \\
    Act-MGN       & 6.4742 & 1.7099 & 1.4075 & 0.0427 \\
    \textbf{Ours} & 2.8659 & 0.6431 & 0.9182 & 0.0339 \\
    \bottomrule
  \end{tabular}
\end{table}

\begin{table}[tb]
  \centering
  \caption{Runtime comparison at
    different batch sizes. Times are per batched step in
    milliseconds. MJX and Act-FIGNet run on the same GPU,
  MuJoCo on 8 CPU cores.}
  \label{tab:runtime}
  \footnotesize
  \setlength{\tabcolsep}{3pt}
  \ra{1.2}
  \begin{tabular}{@{}l *{6}{S[table-format=3.2, round-mode=places,
    round-precision=2]}@{}}
    \toprule
    \multirow{2}{*}{Model} & \multicolumn{6}{c}{Batch size $B$} \\
    \cmidrule(lr){2-7}
    & {1} & {4} & {16} & {64} & {256} & {512} \\
    \midrule
    MuJoCo        & 0.99 & 1.53 & 4.27 & 16.10 & 60.00 & 113.09 \\
    MJX           & 50.08 & 95.04 & 177.09 & 196.99 & 294.99 & 369.08 \\
    \textbf{Ours} & 10.83 & 11.36 & 13.37 & 23.28 & 68.20 & 125.84 \\
    \bottomrule
  \end{tabular}
\end{table}
\subsection{Model analysis}
\label{sec:model_analysis}
\begin{figure}[tb]
  \centering
  \includegraphics[width=0.99\linewidth]{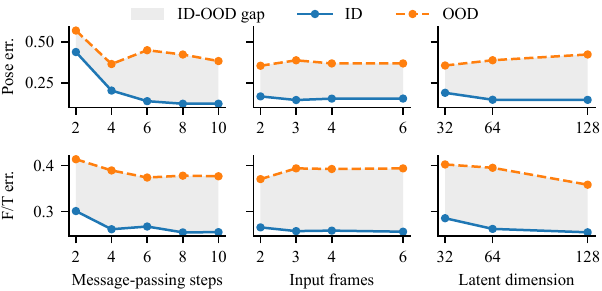}
  \caption{Sensitivity sweep. Errors are normalized by constants
    taken from the ID data: the mean cumulative traveled distance and
    rotation of the GT for pose, and the mean absolute GT readings for
  F/T. The OOD curves are pooled over the three OOD policies.}
  \label{fig:sensitivity_sweep}
\end{figure}

\paragraph{Runtime}We compare per-step runtime of Act-FIGNet, MuJoCo
and MJX over batch sizes in Table~\ref{tab:runtime}, benchmarked on
the same scene and mesh resolution. Our model is consistently faster
than MJX and slower than the CPU-optimized MuJoCo, though that gap
narrows to 1.1$\times$ at $B=512$. Neither GPU model is fast enough
for real-time MPC control; closing that gap requires optimizing the
model jointly with the planner, as in
MuJoCo-MPC~\cite{journals/corr/abs-2212-00541}. In contrast,
sampling-based filtering is far less demanding. With 256 particles,
our models runs under 70\,ms, making online state estimation feasible.
\paragraph{Sensitivity analysis}
A sensitivity sweep on message-passing steps, input frames and latent
dimension is reported in Fig.~\ref{fig:sensitivity_sweep}. Metrics
are normalized pose and F/T errors under random (ID) and OOD
policies. Both ID metrics drop with increasing MP-steps, saturating
at 6. Input frames have little effect while a larger latent dimension
improves ID metrics but degrades OOD pose accuracy.
\paragraph{Antisymmetric bilinear channel}
The channel is added to the object-mesh edge functions
$\phi^{\text{proc}_{l}}_{\mathcal{E}^{O \leftrightarrow M}}$. The
edge function input $\bm{z}$ is first mapped into $\bm{W}_{a}\bm{z}$
and $\bm{W}_{b}\bm{z}$, which are then reshaped into $\bm{A}, \bm{B}
\in \mathbb{R}^{n \times d}$; their antisymmetric products within
each channel are collected:
\begin{equation*}
  \bm{\psi}^{\wedge} = \bigl[A_{k,i}B_{c,j} - A_{k,j}B_{k,i}\bigr]_{
  k \leqslant n,i<j\leqslant d}
\end{equation*}
alongside a symmetric term $\bm{\psi}^{\odot} = (\bm{W}_{p}\bm{z})
\odot (\bm{W}_{q}\bm{z})$ carrying the elementwise products. The edge
function input then becomes $[\bm{z},\bm{\psi}^{\wedge},\bm{\psi}^{\odot}]$.
Since the F/T errors are nearly unchanged, we report the channel's contribution as 100-step pose RMSE in
Table~\ref{tab:bilinear}. The OOD metrics improve by 4 to 26\% in position and by 15 to 63\% in orientation, while the ID metrics move little.
\begin{table}[!tb]
  \centering
  \caption{Antisymmetric bilinear channel ablation. 100-step pose
  RMSE under the ID and OOD policies.}
  \label{tab:bilinear}
  \footnotesize
  \setlength{\tabcolsep}{3.5pt}
  \ra{1.2}
  \begin{tabular}{@{}l
      S[table-format=2.2, round-precision=2]
      S[table-format=2.2, round-precision=2]
      S[table-format=2.2, round-precision=2]
      S[table-format=2.2, round-precision=2]
      S[table-format=1.2, round-precision=2]
      S[table-format=1.2, round-precision=2]
      S[table-format=1.2, round-precision=2]
    S[table-format=1.2, round-precision=2]@{}}
    \toprule
    \multirow{2}{*}{Channel}
    & \multicolumn{4}{c}{Position\,(\%)} &
    \multicolumn{4}{c}{Orientation\,(deg)} \\
    \cmidrule(lr){2-5} \cmidrule(lr){6-9}
    & {Rand.}  & {Circ.\,l} & {Circ.\,h} & {Exp.} & {Rand.} &
    {Circ.\,l} & {Circ.\,h} & {Exp.}\\
    \midrule
    Off & 5.2354 & 13.9610 & 12.9256 & 22.2885 & 0.6717 & 1.8198 &
    1.1887 & 3.0589 \\
    On  & 5.2634 & 10.3684 & 10.9848 & 21.4588 & 0.6414 & 0.6782 &
    0.8546 & 2.5925 \\
    \bottomrule
  \end{tabular}
\end{table}
\section{CONCLUSIONS}
We presented an action-conditional learnable forward and observation
model for peg-in-hole tasks. The model is trained
self-supervised on robot joint and force-torque readings only. The simulation
experiment shows that our approach can be deployed in model-based
control for peg insertion with unseen concave geometries and action
distributions, and that fine-tuning on its own rollouts brings the
success rate to that of an agent planning with the ground truth
dynamics at the tightest clearance. On our real-world setup, our
model shows its advantage over the analytical ones in capturing
realistic contact dynamics, predicting motion and F/T readings more
accurately. The precise F/T prediction points to a role as an
observation model in state estimation.
Our evaluation varies the tool geometry and the action distribution,
but keeps the
contact material and the robot fixed. Generalization across friction,
stiffness changes
and across platforms with different controller compliance remains open.
Future work will integrate predictive uncertainty for state estimation.

\section*{APPENDIX}

\subsection{MPC agent details}
\label{app:mpc_details}
\paragraph{Reward function}
The tip-target displacement $\Delta\bm{p}_T$ is decomposed into an axial
component $\Delta z_T$ and a lateral component $\Delta d_T$, and the
two terms of
$r_T$ are
\begin{gather*}
  \rho(\Delta\bm{p}_T) = \Big[e^{-\Delta z_T/\sigma_z}
  + w_z\big(1 - \min(\tfrac{\Delta z_T}{z_\mathrm{span}},\,1)\big) \Big]
  e^{-\Delta d_T/\sigma_d} \\
  \tau(\Delta\theta_T) = w_\theta\big(\sin^2\!\Delta\theta_T -
  \sin^2\!\theta_c\big)_{+}
\end{gather*}
with $(\cdot)_{+}=\max(\cdot,0)$. The lateral gate $e^{-\Delta d_T/\sigma_d}$
prevents depth progress from being rewarded beside the slot, and $\tau$ leaves
tilts within $\theta_c$ unpenalized. We use $\sigma_z=0.1$, $\sigma_d=0.025$,
$z_\mathrm{span}=0.12$, $w_z=5.0$, $w_\theta=20.0$ and $\theta_c=\ang{2.0}$.
\paragraph{MPC agent}
The agent uses a planning horizon of 12 and replans every step.
Implementation is adapted from
MBRL-lib\cite{journals/corr/abs-2104-10159}, using iCEM as the
optimizer\cite{journals/corr/abs-2008-06389} with 160 samples and 8
optimization iterations.
\subsection{System identification}
\label{app:baseline_details}
The full robot model equipped with a closed-loop force controller is
fit in two stages: collision-free trajectories set the controller
gains, a global joint-dynamics scale, and the tool mass to minimize
pose error; these are then frozen while the socket and tool contact
solver parameters are fit on data with contact minimizing both
trajectory and wrench errors. The simplified model has no controller,
so its parameters for dynamics and contact are jointly fit in a
single stage on contact-rich data against pose and wrench errors.

\section*{ACKNOWLEDGMENT}
The DFKI Niedersachsen is sponsored by the Ministry of
Science and Culture of Lower Saxony and the Volkswagen Stiftung. This
work was funded through a grant of the German Federal Ministry for
Economic Affairs and Climate Action with the grant number 01ME19003D.
We thank Yulia Rubanova for her valuable advice on the FIGNet model
implementation.

\balance
\bibliographystyle{IEEEtran}
\bibliography{refs}

\clearpage
\section*{SUPPLEMENTARY MATERIAL}
\appendices
\subsection{Graph construction details}
\label{app:graph_construction_details}
\subsubsection{Node features}
\paragraph{Mesh node}Mesh node features are concatenation of vertex
velocities and the object's static attributes. Vertex velocities are
estimated by
the difference between positions in two consecutive frames:
\begin{equation}
  \bm{v}^{M, \text{feature}}_i = [\bm{p}^{m_i}_t -
  \bm{p}^{m_i}_{t-1}, ..., \bm{p}^{m_i}_{t-h+1}-\bm{p}^{m_i}_{t-h}, \bm{k}]
\end{equation}
where $\bm{p}^{m_i}_t \in \mathbb{R}^3$ is the vertex position, with the
superscript, subscript $(\cdot)^{m_i}$, $(\cdot)_t$ indicating the vertex and
time index respectively. Here, $\bm{k} = [m, \mu, b]$ represents the
mass, friction coefficient
and a binary variable related to its dynamic state. Object node features are
computed similarly, but
with object linear and angular velocities. In the experiments, $\mu$
and $m$ are constants; we set $h=3$ in the simulation and set to 2 in
real-world experiment. Larger $h$ slightly degrades OOD performance
shown in Fig.~\ref{fig:sensitivity_sweep}
\paragraph{Wrench node} Their features are simply one-hot vectors:
\begin{equation}
  \bm{v}^{W, \text{feature}}_{i}=[0, 1, 0]
\end{equation}
When controller parameters vary, it offers a place to condition on.
\subsubsection{Edge features}
\paragraph{Wrench-mesh edge}\label{app:world_mesh_edge}The features
are per-vertex coupled force field $\{\bm{f}^{m_i}\}$ calculated
according to \eqref{eq:force_edge_feature}, and concatenated with its magnitude:
\begin{gather}
  \bm{e}^{\text{feature}}_{w \to m_i} =
  \bigl[\bm{f}^{m_i},\;\|\bm{f}^{m_i}\|\bigr]
\end{gather}
\paragraph{Mesh-mesh edge}\label{app:mesh_mesh_edge}
In contrast to the face-face edges in\cite{conf/iclr/AllenRL0SBP23}, we
directly create edges between mesh node triplets belonging to the
potentially colliding faces.
First, the
collision detection algorithm provides the closest points on a
colliding face pair, denoted as
$\bm{p}^s$ and $\bm{p}^r$. Next, the vertices of each face are ranked
by their distances to the respective closest point, resulting in
ordered vertex sets
$\{\bm{p}^{m_{s_i}}\}_{i=1\dots3}$ and $\{\bm{p}^{m_{r_i}}\}_{i=1\dots3}$. In
the following, we use $m_{\bm{s}}$ and $m_{\bm{r}}$ to denote the ordered vertex
indices, i.e., $m_{\bm{s}/\bm{r}}=\{m_{{s_i/r_i}}\}_{i=1\dots3}$.
Features are first projected to the receiver's body frame then concatenated:
\begin{equation}
  \bm{e}^{\text{feature}}_{m_{\bm{s}} \to m_{\bm{r}}} =
  \bigl[\bm{d}_{rs}, [\bm{d}_{s_i}]_{i=1\dots3},
  [\bm{d}_{r_i}]_{i=1\dots3}, \bm{n}_s, \bm{n}_r\bigr]
\end{equation}
where $\bm{d}_{s_i/r_i} = \bm{p}^{m_{s_i/r_i}}_t - \bm{p}^{s/r}$ are the vectors
from face vertices to their corresponding closest points;
$\bm{d}_{rs}$ is the vector from $\bm{p}_s$ to
$\bm{p}_r$; and $\bm{n}_s$ and $\bm{n}_r$ the normals of both faces.
Message-passing
for mesh-mesh connections is handled differently: edge updates take
into account all six
sender and receiver nodes:
\begin{gather}
  \bm{e}^{l+1}_{m_{\bm{s}} \to m_{\bm{r}}} =
  \phi^{\text{proc}_{l}}_{\mathcal{E}^{M \to
  M}}\bigl(\bigl[\bm{e}^{l}_{m_{\bm{s}} \to m_{\bm{r}}}, \bm{v}^{M,
  l}_{\bm{s}}, \bm{v}^{M, l}_{\bm{r}}\bigr]\bigr)\\[1ex]
  \bm{v}^{M, l}_{\bm{s}} = [\bm{v}^{M, l}_{s_i}]_{i=1\dots3} \quad
  \bm{v}^{M, l}_{\bm{r}} = [\bm{v}^{M, l}_{r_i}]_{i=1\dots3}
\end{gather}
Each updated edge feature is split into three vectors, one for each
receiver node. Messages from face-face
interaction are then aggregated for each mesh node $m_j$ by summing
up all incoming edge features:
\begin{equation}
  \bm{h}^{m_j} = \sum_{\forall m_{r_i} =
  m_j}{\bm{e}^{l+1}_{m_{\bm{s}} \to m_{r_i}}}
\end{equation}
Finally, the aggregated face-face messages are concatenated with the other
aggregation to update the mesh node features, as shown
in~\eqref{eq:node_update}.

\subsection{Model details}
\label{app:model_architecture_details}
The functions in the EPD stack, including encoders, decoders, node and edge
functions are implemented as two-layer MLPs with ReLU activations. A
LayerNorm module is applied after each MLP, except in the decoders.
Residual connection is used between two message-passing layers.
The implementation is built on
PyTorch\cite{DBLP:conf/asplos/AnselYHGJVBBBBC24} and
PyG\cite{DBLP:journals/corr/abs-2507-16991}. See Table~\ref{tab:epd_params} for
model parameters. We also experimented with
two different threshold values for collision detection, 1mm
and 10mm, and found that larger collision radius slightly degrades
prediction accuracy on the real-world data from our setup.
\begin{table}[b]
  \centering
  \caption{EPD stack parameters}
  \label{tab:epd_params}
  \begin{tabular}{l r l r}
    \toprule
    Parameter & Value & Parameter & Value\\
    \midrule
    Message passing steps & 6 & MLP hidden dimension & 64 \\
    Edge feature dimension & 64 & MLP layers & 2 \\
    Node feature dimension & 64 & Wedge product channel & 2 \\
    Bilinear product rank & 2 & Wedge/bilinear dimension & 6 \\
    \bottomrule
  \end{tabular}
\end{table}

\subsection{Control policies}
\label{app:control_policies}
\paragraph{Random policy for training} We adopt the random trajectory
generation method described in~\cite{conf/icml/Sanchez-Gonzalez18}.
First, random points are sampled uniformly within
action limits, where the number of points corresponds to the episode
length. An action sequence is then generated via cubic spline
interpolation between
these points. Safe execution is ensured by a bounded workspace for
the end effector. Exceeding these limits triggers a virtual spring
force that pulls the end effector back into the valid workspace.
\paragraph{Peg insertion policy} The tool initially descends to
establish surface contact, after which the controller executes a
spiral search trajectory in the $xy$-plane centered at the target
(the bottom of the slot). The spiral motion is coupled with a constant
downward force to ensure surface contact.
\paragraph{Circular trajectory policy} The tool first moves down to
establish contact with
the surface, then follows circular paths around the hole's perimeter.
We test with two
different velocities to evaluate the model's robustness to action
frequency change. In low velocity, the tool completes one full circle in 200
steps (20s), while in high velocity, it completes three circles in
the same number of
steps.

\begin{figure}[tb]
  \centering
  \begin{minipage}{0.99\linewidth}
    \begin{subfigure}[b]{0.99\linewidth}
      \centering
      \includegraphics[width=0.99\linewidth]{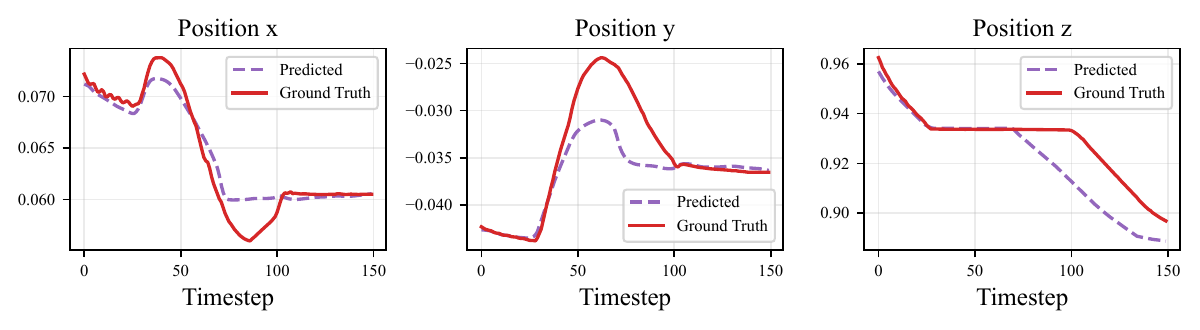}
      \caption{}
      \label{fig:forward_only}
    \end{subfigure}
    \vfill
    \begin{subfigure}[b]{0.99\linewidth}
      \centering
      \includegraphics[width=0.99\linewidth]{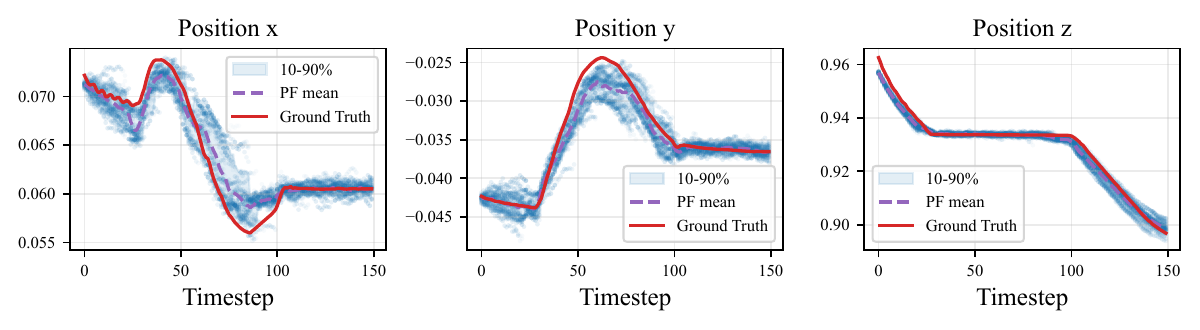}
      \caption{}
      \label{fig:pf_no_init_noise}
    \end{subfigure}%
    \vfill
    \begin{subfigure}[b]{0.99\linewidth}
      \centering
      \includegraphics[width=0.99\linewidth]{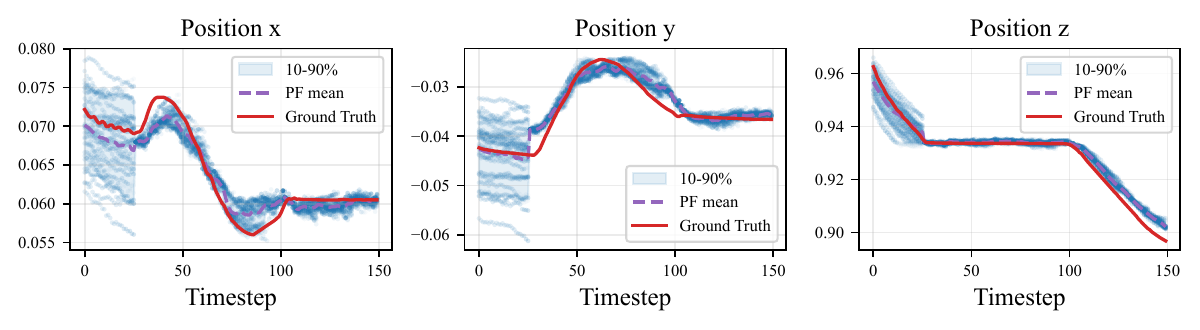}
      \caption{}
      \label{fig:pf_with_init_noise}
    \end{subfigure}
  \end{minipage}
  \caption{Peg position estimation during a real-world peg insertion episode
    a using particle filter with F/T data. (a) open-loop prediction using
    dynamics model only. (b) Trajectory estimation using the particle filter
    with F/T correction, initialized at the ground-truth pose.
    (c) The same filter with F/T correction, but initialized with a random,
    inaccurate pose to simulate noisy perception.
  }
  \label{fig:particle_filter}
\end{figure}
\begin{figure}[b]
  \centering
  \includegraphics[width=0.99\linewidth]{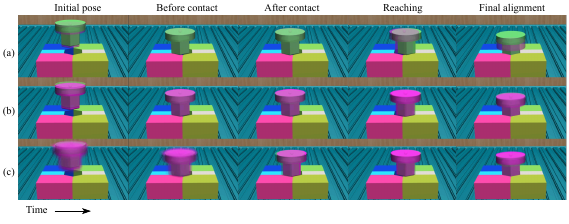}
  \caption{Rendered images of the same real-world peg insertion episode shown in
    Fig.~\ref{fig:particle_filter}, overlaid
    with the particle filter estimates. The green tools represent ground
    truth, while the transparent magenta overlays represent the particle
    estimates/motion prediction. (a) Open-loop prediction. (b) Particle filter
    with accurate initialization. (c) Particle filter with noisy initial pose.
    The rendered video is included in the supplementary video. The magenta
    prediction in (a) is almost invisible before making contact because it
    closely follows the ground truth. In (b)-(c), due to the
    concentration of particles around the ground truth, the magenta overlays
    appear as a single solid body occluding the green ground truth.
  }
  \label{fig:pf_vid}
\end{figure}
\subsection{State estimation with F/T readings} In the simulation
experiments, we assumed perfect knowledge of the peg's pose relative to the
slot. In practice, this information needs to be estimated from sensors,
such as camera, which is often corrupted by noise and subject to visual
occlusion. Furthermore, as previously discussed, open-loop forward
dynamics models suffer from
accumulated errors, leading to significant drift over time. To
address these, we can leverage the
accurate one-step predictive models within a filtering framework to estimate
the relative peg pose online, using F/T sensor data as observations.

In this experiment, we implement a simple particle filter that uses the
learned dynamics model for motion updates, and the learned observation model for
F/T measurement updates. The filter is configured with 30
particles and empirically selected
noise parameters. To evaluate the contribution of each component, we tested
three configurations. (1)\,\textbf{Motion-only}: open-loop prediction using only
the learned dynamics model; (2)\,\textbf{particle filter with known pose}:
a particle filter initialized with the ground truth pose. Estimates
are corrected using F/T measurement updates following the motion
updates.; (3)\,\textbf{particle filter with perturbation}:
identical to (2) but the initial pose is corrupted with significant
error, simulating an inaccurate initial estimate from a vision system.

The resulting estimation trajectories are
illustrated in Fig.~\ref{fig:particle_filter} to
Fig.~\ref{fig:pf_vid}. A comparison between
Fig.~\ref{fig:forward_only} and Fig.~\ref{fig:pf_no_init_noise}
reveals how the F/T observation model effectively
corrects the motion model's drift; even with noisy initial pose
(Fig.~\ref{fig:pf_with_init_noise}). As expected, the
estimates rapidly converge to the ground truth upon making contact, especially
along the $z$-axis, where the F/T observations are the most
informative. More importantly, both the predicted (from the forward
model) and estimated (from the filter) trajectories are physically
plausible, with no interpenetration between objects, because the
motion updates themselves are physically feasible. This is especially
apparent along the $z$-axis, where the predicted and estimated values
consistently remain above the ground truth. This
demonstrates the potential of our learned models to complement visual data for
state estimation in contact-rich manipulation tasks.
\begin{figure}[tb]
  \centering
  \begin{subfigure}[b]{0.3\linewidth}
    \centering
    \includegraphics[width=0.99\linewidth]{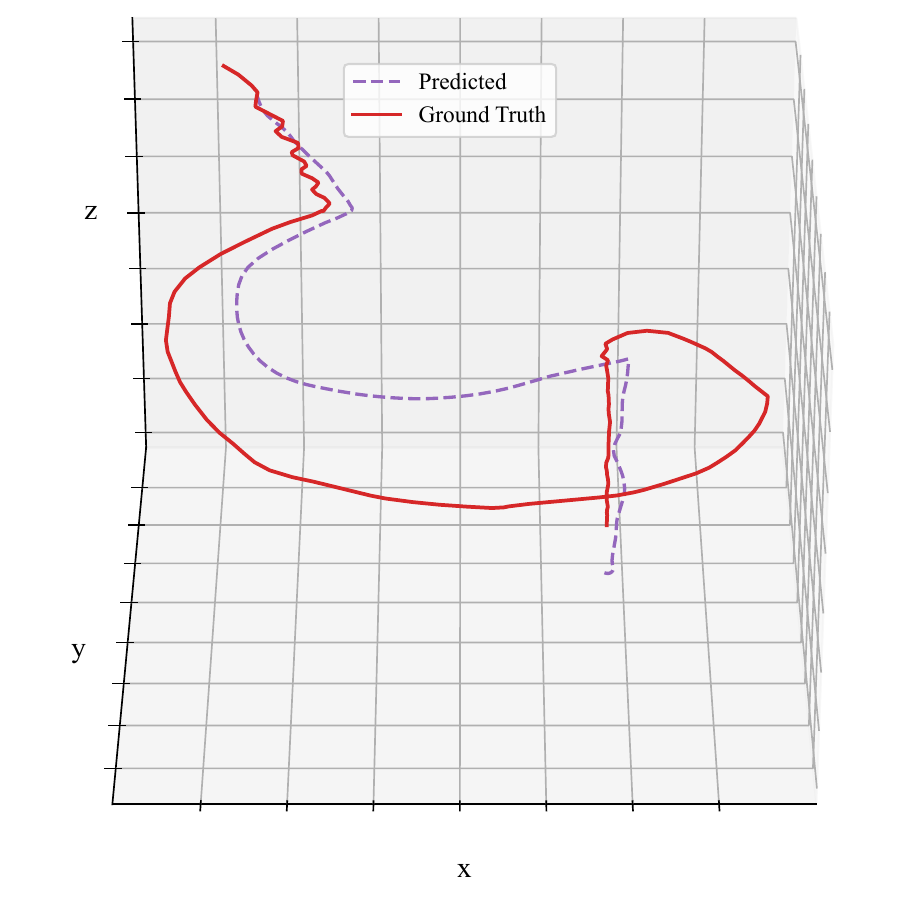}
    \caption{}
    \label{fig:forward_only_3d}
  \end{subfigure}
  \hfill
  \begin{subfigure}[b]{0.3\linewidth}
    \centering
    \includegraphics[width=0.99\linewidth]{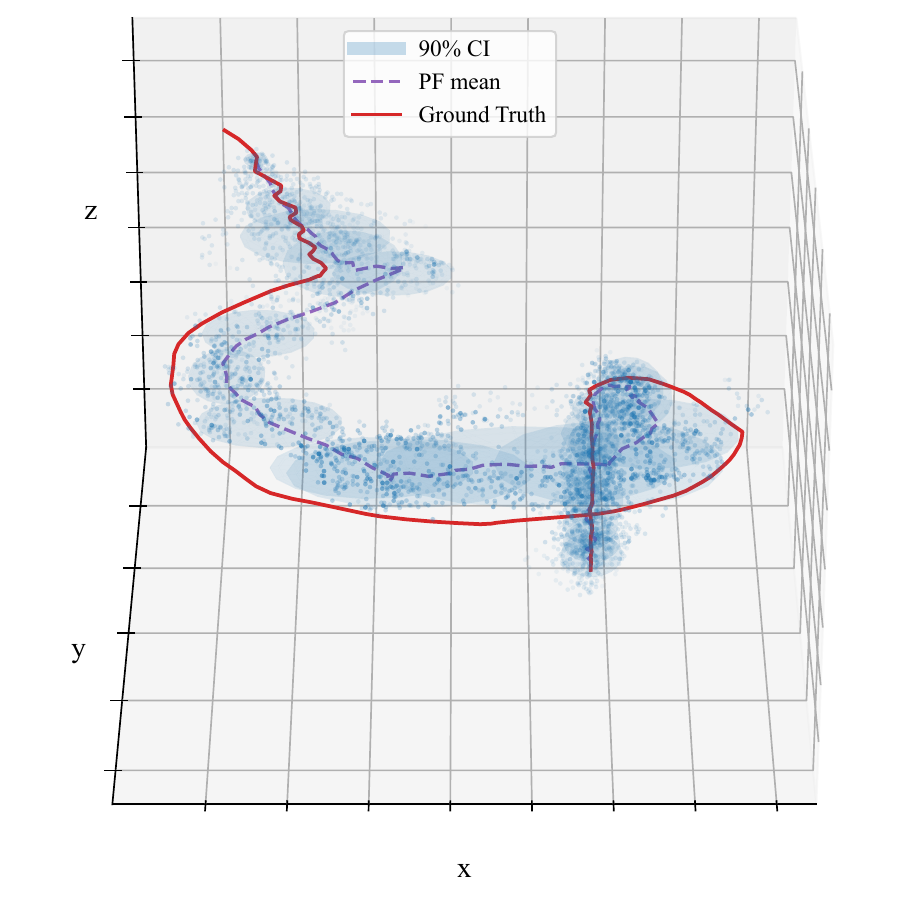}
    \caption{}
    \label{fig:pf_no_init_noise_3d}
  \end{subfigure}%
  \hfill
  \begin{subfigure}[b]{0.3\linewidth}
    \centering
    \includegraphics[width=0.99\linewidth]{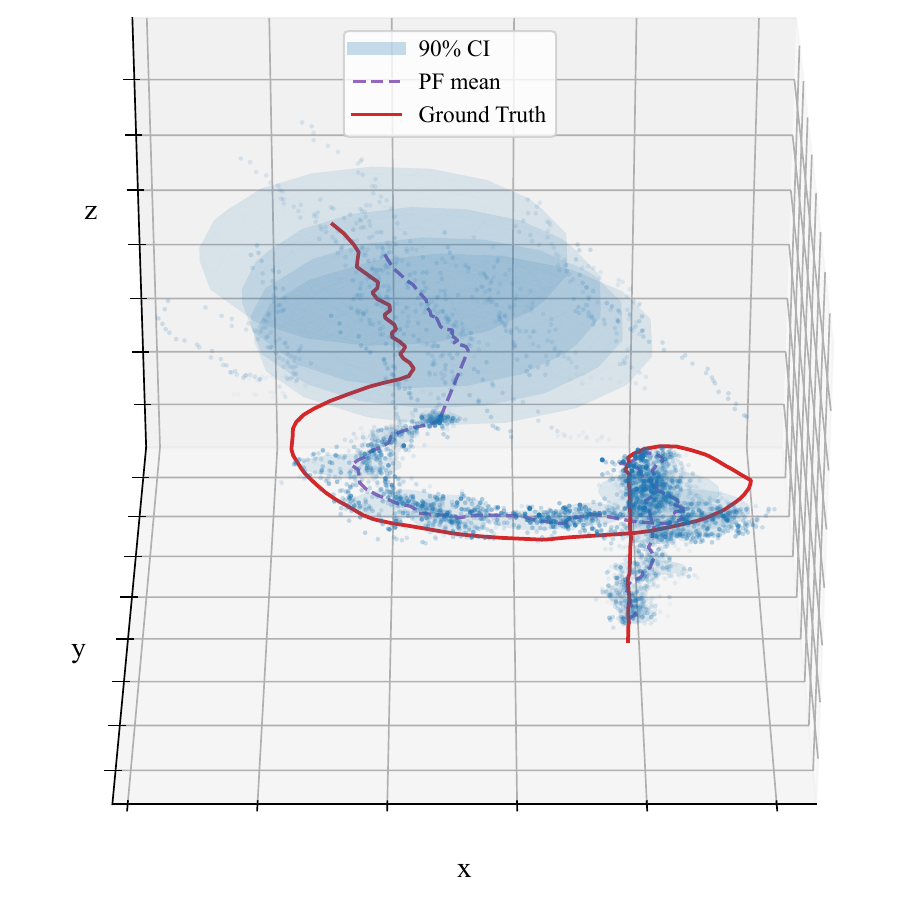}
    \caption{}
    \label{fig:pf_with_init_noise_3d}
  \end{subfigure}
  \caption{3D visualization of a sample peg insertion trajectory,
    complementing the 2D projection shown in Fig.~\ref{fig:particle_filter}.
    (a) Open-loop
    prediction. (b)
    Particle filter
    with accurate initialization. (c) Particle filter with noisy initial pose.
  }
  \label{fig:particle_filter_3d}
\end{figure}

\end{document}